\documentclass[]{interact}

\usepackage{enumitem}
\usepackage{graphicx}
\usepackage{xpatch}
\usepackage{hyperref}
\usepackage{setspace}
\usepackage{footmisc}
\usepackage{multirow}
\usepackage{tabularray}
\usepackage{colortbl}
\usepackage{color}
\usepackage{array}
\usepackage{tcolorbox}
\usepackage{float, url}
\usepackage{CJKutf8}
\usepackage{adjustbox}
\usepackage{makecell}
\usepackage{threeparttable}
\usepackage{epstopdf}
\usepackage{subfigure}
\usepackage{natbib}
\bibpunct[, ]{(}{)}{;}{a}{}{,}

\theoremstyle{plain}

\theoremstyle{definition}

\theoremstyle{remark}

\hypersetup{
	colorlinks,
	linkcolor={orange!95!black},
	citecolor={blue!90!black},
	urlcolor={olive!95!black}
}

\usepackage{imakeidx}
\makeindex[columns=1, intoc]

\begin{document}

\articletype{}

\title{\textbf{Beyond Traditional Algorithms: Leveraging LLMs for Accurate Cross-Border Entity Identification}
\footnote{\textbf{\textit{Acknowledgments}}: We thank Carmen Astorga, Eugenia Chamorro, Jose Manuel Carbó and Carlos García Fernandez for valuable feedback and suggestions.}}

\author {
\name{Andres Azqueta-Gavaldon\textsuperscript{$\dagger$}, Joaquin Ramos Cosgrove\textsuperscript{$\dagger$}} 
\textsuperscript{$\dagger$}Financial Information \& Central Credit Register Department, Banco de España \\
}

\maketitle

\bigskip
\bigskip

\begin{abstract}

The growing prevalence of cross-border financial activities in global markets has underscored the necessity of accurately identifying and classifying foreign entities. This practice is essential within the Spanish financial system for ensuring robust risk management, regulatory adherence, and the prevention of financial misconduct. This process involves a labor-intensive entity-matching task, where entities need to be validated against available reference sources. Challenges arise from linguistic variations, special characters, outdated names, and changes in legal forms, complicating traditional matching algorithms like Jaccard, cosine, and Levenshtein distances. These methods struggle with contextual nuances and semantic relationships, leading to mismatches. To address these limitations, we explore Large Language Models (LLMs) as a flexible alternative. LLMs leverage extensive training to interpret context, handle abbreviations, and adapt to legal transitions. We evaluate traditional methods, Hugging Face-based LLMs, and interface-based LLMs (e.g., Microsoft Copilot, Alibaba’s Qwen 2.5) using a dataset of 65 Portuguese company cases. Results show traditional methods achieve accuracies over 92\% but suffer high false positive rates (20–40\%). Interface-based LLMs outperform, achieving accuracies above 93\%, F1 scores exceeding 96\%, and lower false positives (40–80\%).
\end{abstract}

\begin{keywords}
Large language models; natural language processing; machine classification; entity matching.
\end{keywords}

\setlength{\parskip}{0.5em}

\clearpage{}

\tableofcontents

\clearpage{}

\section{\index{Introduction}Introduction}\label{sec:Introduction}

The correct identification of foreign entities reported by credit institutions is essential for an effective control, monitoring and management of financial risks. This process involves assigning a unique identification code which is key to maintaining operations tracking and risk valuation at an optimal level. In the current global market context, an accurate identification of foreign entities also helps regulatory authorities to better monitor credit institutions’ economic and financial activities, reinforcing national and international standards compliance as well as financial information transparency and integrity. Additionally, these unique identifications play a critical role in preventing fraud and money laundering by providing a standardized method for the identification of counterparties involved in financial transactions.


These identifications are currently assigned through a labor-intensive entity-matching process which consists of receiving a daily list of foreign entities whose details (name, address, legal form...) are compared against the available source of reference (hereinafter referred to as ASR). ASR includes a series of datasets sourced from a wide range of different national and international databases such as Einforma (mainly Spain and Portugal), Companies House (UK) or Bundesanzeiger (Germany). If the information of the incoming record matches all attributes in the ASR, the identification will be approved and given a unique code (a new or an existing one). On the contrary, if there is no match (or a very poor matching) between the incoming data and the ASR, the incoming record will be rejected. Therefore, there is a permanent entity-matching challenge since small differences between incoming data and the ASR could easily lead to wrong conclusions, for example, considering two datasets as different entities when they are actually referring to the same one and vice versa. Finally, there are a few other challenging elements that can add a significant level of difficulty to the already complex identification process:  language-specific character variations such as ô, ł, ç, š, ò; use of outdated names\footnote{A company may have changed its name, but the old name is still recorded. In such cases, the entity’s code remains unchanged, even though the name no longer matches.} or legal form changes\footnote{This occurs when a company changes from one legal structure to another, such as from a limited liability company to a partnership. Although the legal form changes, the entity itself remains the same, and its code should stay consistent.}, among others.

Given the complex nature of this entity-matching problem, traditional matching algorithms such as Jaccard, cosine, and Levenshtein distances are limited. These algorithms rely heavily on string similarity or token overlap, which can fail to account for contextual nuances. For instance, Jaccard and cosine similarity may struggle with entities that have minor typographical differences, such as 'International Corp.' versus 'Internat’l Corp.' or 'Müller AG' versus 'Mueller AG,' where special characters or abbreviations alter the textual representation without changing the entity's identity. Similarly, Levenshtein distance, which measures the number of edits needed to transform one string into another, may incorrectly flag entities with slight variations as entirely different. For example, 'Tech Solutions Ltd.' and 'Tech Sols Limited' might be deemed dissimilar despite referring to the same company. Furthermore, these algorithms do not inherently account for semantic relationships, such as legal name changes ('OldCo Inc.' to 'NewCo LLC') or structural differences in legal forms (e.g., 'ABC GmbH' versus 'ABC KG'). As a result, traditional methods often fall short in scenarios requiring deeper contextual understanding or domain-specific knowledge. Given that this work focuses on Portugal, the legal forms considered are the Sociedade Limitada (Lda.), Sociedade Anónima (SA), and Sociedade Civil (SC). Additionally, linguistic particularities such as the characters 'ç', 'ã', and 'õ'—which do not exist in Spanish—are also taken into account.

In this work, we explore the use of Large Language Models (LLMs) as a more flexible and adaptive alternative to traditional matching algorithms. Unlike rule-based or distance-based methods, LLMs can leverage their extensive training on diverse textual data to understand context, account for semantic relationships, and handle variations in entity representation. For instance, LLMs can recognize that 'Tech Solutions Ltd.' and 'Tech Sols Limited' likely refer to the same entity by interpreting abbreviations, synonyms, and contextual cues. Additionally, LLMs can accommodate language-specific nuances, such as special characters or transliterations, and adapt to domain-specific knowledge, such as legal changes in name or legal form. By incorporating these capabilities, LLMs offer a promising approach to overcoming the limitations of traditional methods, enabling more accurate and robust entity matching in complex scenarios.

To evaluate these wide range of algorithms, it is essential to establish a dataset with a well-defined ground truth. Specifically, the dataset must clearly indicate whether two entities are the same or not, providing definitive "accept" or "reject" labels for each pair. To achieve this, we constructed a database using legal entity data from Portugal. This choice was motivated by two key factors. First, the linguistic similarities between Portuguese and Spanish allow us to leverage bilingual expertise during the labeling process, thereby minimizing human errors. Second, open-source resources such as \textit{Informa} are available in Portugal, enabling us to cross-check and validate the data more effectively. In contrast, many other countries lack well-established, openly accessible firm registries, making it significantly more challenging to construct a reliable and verifiable dataset. 

The resulting dataset consists of 65 well-defined cases, each including detailed information on legal forms and various identification codes. Additionally, we include diverse name variations, such as abbreviations, omitted punctuation, and misspellings, allowing us to evaluate the strengths and weaknesses of the different models under consideration. We categorize the models into three groups: traditional models, which include distance-based metrics such as Cosine similarity, Levenshtein distance, and Jaccard similarity; LLMs operated via the Hugging Face ecosystem, a widely adopted open-source platform that provides tools, libraries, and pre-trained models for natural language processing (NLP); and LLMs accessed via interfaces, which are open-source LLMs that can be run through user-friendly interfaces, including Microsoft Copilot, Qwen 2.5 from Alibaba, and the Mistral model.

Our analysis reveals significant variations in performance across traditional distance-based methods and state-of-the-art Large Language Models (LLMs) in the context of entity matching (EM). Traditional methods, such as Levenshtein and Cosine similarity, demonstrated robust performance with accuracies exceeding 92\% and strong precision and recall metrics. However, these methods exhibited notable limitations in minimizing false positives, with False Positive Rates (FPR) ranging from 20\% to 40\%. Among LLMs, deBERTa-v3-base achieved perfect recall and a high F1 Score of 95.87\%, but its inability to distinguish classes effectively (ROC AUC of 50\%) and an FPR of 100\% highlighted challenges in avoiding false positives. In contrast, interface-based LLMs like Microsoft/Copilot and Alibaba's Qwen2.5 showcased exceptional performance, achieving accuracies above 93\%, F1 Scores exceeding 96\%, and more acceptable FPRs of 40–80\%. These findings underscore the superior capabilities of interface-based LLMs in handling complex classification tasks, likely due to their extensive training on diverse datasets and advanced architectures that capture nuanced contextual relationships. However, challenges remain in scenarios involving abbreviations or lexical variations, where models struggle due to insufficient fine-tuning on domain-specific patterns. Overall, our study highlights the potential of LLMs to enhance EM processes by improving accuracy and efficiency while minimizing human intervention. Yet, given the critical need for zero tolerance in accepting incorrect entities (false positives) —especially in legal and financial contexts—further refinements are necessary to reduce FPR and ensure consistent performance across diverse datasets. This work lays the foundation for developing more reliable and scalable EM solutions.

The structure of the paper is organized as follows. Section 2 presents the literature review, focusing on entity matching and Large Language Models (LLMs). The subsequent section, Section 3, outlines the workflow of our proposed application, detailing the data and the various methods considered. Section 4 introduces the use case of Portugal, where we constructed our dataset to perform the analysis. The following section, Section 5, presents the results, while the final section provides conclusions and suggestions for future research.


\section{\index{Literature Review}Literature Review}\label{sec:literature}

Entity matching has proven to be of paramount importance across a wide range of fields. In data integration, it plays a critical role in combining data from multiple sources into a unified view (see \cite{brizan2006survey}, \cite{getoor2012entity}, \cite{li2021deep}). E-commerce platforms such as Amazon, eBay, and Alibaba rely heavily on entity matching to ensure accurate and consistent product listings (\cite{zuo2020flexible}, \cite{tracz2020bert}). Search engines and knowledge graphs utilize entity matching to link queries to specific entities within their knowledge bases, improving search relevance and accuracy (\cite{cheng2007supporting}, \cite{jain2015enhancing}). Social media and online communities also benefit from entity matching (\cite{kumar2013entity}), as do healthcare and biomedical research, where it consolidates patient records, medical research data, and clinical trials (\cite{mccoy2013matching}, \cite{kusa2023effective}). Fraud detection systems leverage entity matching to identify patterns or duplicate identities (\cite{verma2017fraud}), while recommendation systems use it to resolve ambiguities in user-item interactions, thereby enhancing recommendation quality (\cite{he2020mining}). In legal and compliance domains, entity matching ensures accurate and consistent data to meet regulatory requirements (\cite{kruse2021developing}). Similarly, government and public sector applications rely on entity matching to manage citizen records, census data, and public services (\cite{dutta2017census}, \cite{wong2024semantically}). These examples illustrate just a few of the many areas where entity matching has become indispensable.

Despite its widespread utility, entity matching faces several challenges that complicate its implementation. A primary issue is the presence of noisy and incomplete data, which often stems from variations in formatting, abbreviations, or missing attributes (\cite{christen2012data}). For example, names may differ due to typographical errors, transliterations, or cultural differences, making it difficult to determine whether two records refer to the same entity. Scalability is another significant hurdle, as the computational cost of comparing all possible pairs of records grows quadratically with dataset size (\cite{papadakis2023critical}). This challenge is particularly acute in large-scale applications, such as e-commerce platforms or government databases that handle millions of entities. Traditional rule-based approaches, which rely on handcrafted similarity metrics and thresholds, struggle to generalize across diverse domains and languages (\cite{elmagarmid2006duplicate}) and often fail to capture semantic relationships between entities.

To address these limitations, recent advancements in Natural Language Processing (NLP) have introduced powerful machine learning techniques. Transformer-based models, such as BERT and its variants, excel at capturing contextualized representations of text, enabling more accurate comparisons between entity descriptions (\cite{peeters2023entity}, \cite{tracz2020bert}, \cite{li2021deep}). These models understand subtle linguistic nuances, such as synonymy, polysemy, and domain-specific terminology, thereby improving robustness. For instance, Ditto (\cite{li2020deep}) frames entity matching as a sequence-pair classification problem, achieving up to a 29\% increase in F1 score on benchmark datasets compared to traditional methods. Similarly, generative Large Language Models (LLMs) have shown promise in zero-shot and task-specific scenarios, adapting to out-of-distribution entities without extensive retraining (\cite{peeters2023entity}).

Innovative frameworks have further advanced the field. For example, COMEM (\cite{wang2024match}) integrates three strategies—binary classification of record pairs, pairwise comparison, and candidate selection—to balance performance and computational cost. By first filtering candidates using lightweight methods and then applying more intensive techniques for fine-grained identification, COMEM achieves both efficiency and accuracy. Additionally, LLMs demonstrate versatility in handling unstructured data, reducing the need for feature engineering and excelling in low-resource scenarios (\cite{huang2024leveraging}). Recent work also explores unsupervised and weakly supervised methods, leveraging LLMs to generate synthetic data and provide rich contextual embeddings.

While fine-tuning LLMs offers additional opportunities, its effectiveness varies depending on model size and training data (\cite{steiner2024fine}). Structured explanations, for instance, can enhance performance in certain cases, though larger models exhibit mixed results. Despite these advancements, challenges remain, including the need for domain adaptation, performance degradation in low-resource languages, and computational constraints associated with deploying LLMs at scale.

In summary, while traditional entity matching techniques laid the foundation for the field, modern NLP methods—particularly transformer-based models and LLMs—have significantly advanced its capabilities. These innovations address key challenges related to noise, scalability, and semantics, enabling more robust and adaptable solutions. However, ongoing research is needed to overcome remaining obstacles and ensure that entity matching systems remain effective, equitable, and scalable in increasingly complex and diverse settings.

\section{\index{Workflow, Data and Methods}Workflow, Data and Methods}\label{sec:Workflow_Data_Methods}

\subsection{\index{Workflow}Workflow}\label{sec:Workflow}

\begin{figure}[H]
    \centering
    \includegraphics[width=1.2\linewidth]{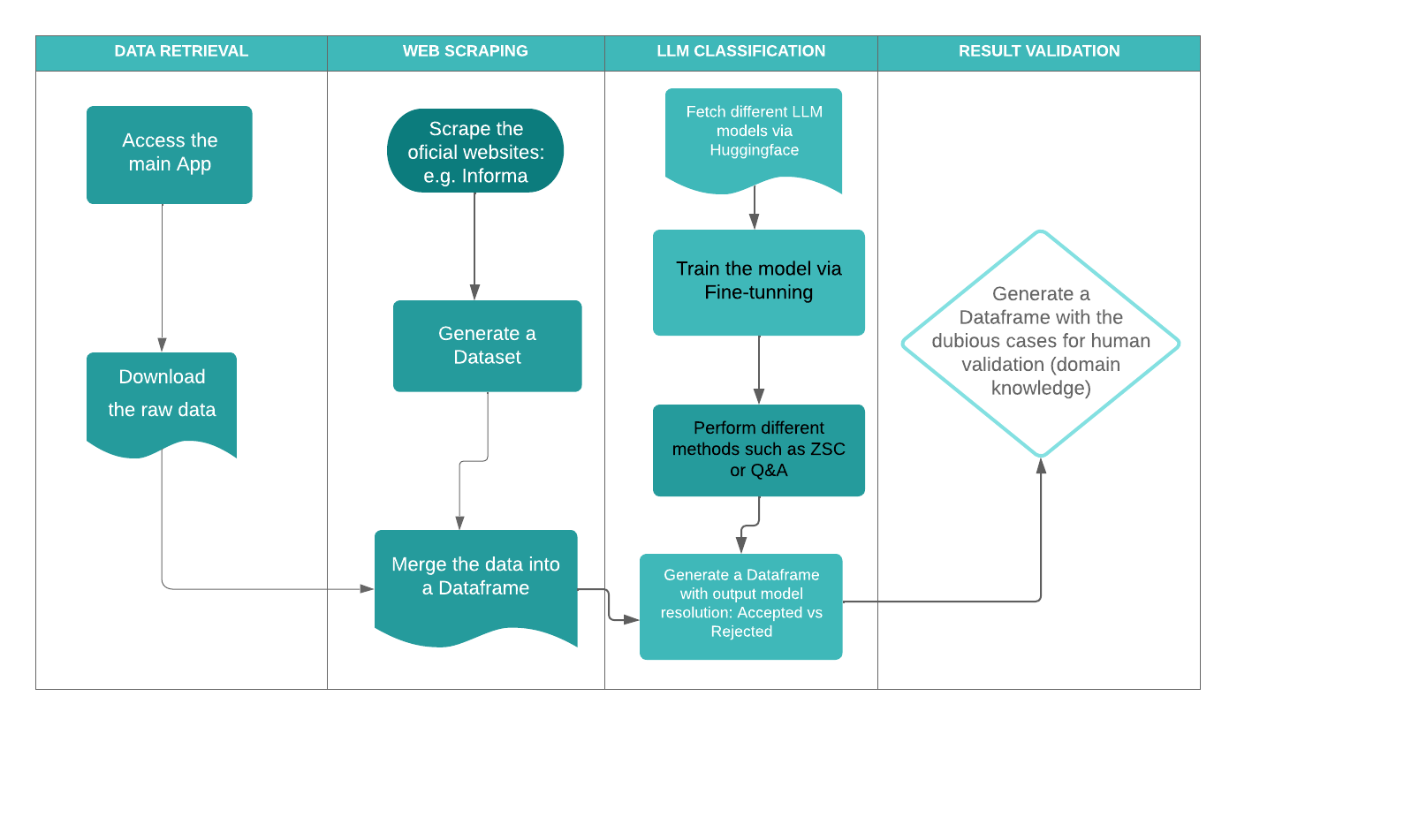}
    \caption{Workflow}
    \label{fig:workflow}
\end{figure}

To address the limitations of the manual comparison process (see Appendix \ref{sec:Manual_Comparison_Process}), an automated comparison process is proposed in this work. As can be seen in Figure \ref{fig:workflow}, the proposed workflow is form of four main pillars: \textit{data retrieval}, \textit{web scrapping of publicly available information}, \textit{LLM classification}, and \textit{result validation}. 

\begin{enumerate}
    \item \textbf{Data retrieval:} The first thing we need to do before starting is to access the main app we use and download the data we are going to use to assess the resolution. Before proceeding further, we compare the incoming foreign entities against the RIAD database to determine whether these entities have already been assigned unique identification codes. This step ensures we avoid redundant processing by reusing existing codes for entities already present in RIAD, thereby reducing computational load and improving accuracy.
    \item \textbf{Web scraping:} Secondly, knowing that our data is not complete, we shall do so. To fill in the official names column, web scraping is done by taking the identifier variable and searching in the official websites (Informa...). These websites usually correspond to national registers which contain all the bureaucratic information concerning each entity.
    \item \textbf{LLM Classification:} At this point in time, we are ready to pass our Dataframe to the LLM to predict the different labels. We shall not forget that the LLM has been previously fine-tuned with data regarding some fictitious entities. In order to the LLM to predict precisely, ZSC or Q\&A method will be used. In both cases the predict labels are: "Accepted" (the information declared matches to the reference available data, RAD), "Rejected (the information declared has some errors, ranging from misspelling to previous changes) and "Doubtful" (this is and extraordinary case when the model is not able to classify correctly the observation, thus to ensure robustness it is preferable to take a closer look).
    \item \textbf{Result validation:} In the end, after all the analysis is done, the cases where the model predicted a "Doubtful" observation will be reviewed meticulously by a specialized human. He will decide wether the entity will be rejected or Accepted depending on the information stated.
\end{enumerate}

\subsection{\index{Data}Data}\label{sec:Data}

\begin{table}[h]
\begin{adjustbox}{width=1\textwidth, center}
    \centering
    \begin{threeparttable}
    \scriptsize
    \begin{tabular}{ccccccccc}
        \hline
        \textbf{Country} & \textbf{Company name} 
        & \textbf{National Identifier} & \textbf{Identifier type} & \textbf{LEI\tnote{a}} 
        & \textbf{Legal Form} & \textbf{Abbreviation}\\
        \hline
        & & \\
        US & \makecell[l]{ISHARES US \\ TRASNPORTATION ETF} 
        &  &  & X 
        & AAXXX & OTROS\\
        & & \\
        RO & \makecell[l]{AIC AUTOLIV ROMANIA \\ S.R.L-IRO DIVISION} 
        & AAXXXX & RO\_CUI\_CD &  
        & AAXXX & S.R.L.\\
        & & \\
        DK & LESJOFORS A/S 
        & XXXX & DK\_CVR\_CD & 
        & AAXXX & A/S\\
        & & \\
        \hline
    \end{tabular}
    \caption{Example Dataset}
    \begin{tablenotes}
    \item[a] Legal Entity Identification
    \end{tablenotes}
    \end{threeparttable}
    \label{tab:BD_ejemplo}
    \end{adjustbox}
\end{table}

The dataset used in this work, contains a detailed overview of information about the entities such as the country of residence, social denomination, sector, or legal form to name a few (see Table \ref{tab:field_descriptions} for all detail information and Table \ref{tab:BD_ejemplo} as an example). Of a particular interest in our case is the information regarding the legal form and social denomination as this information tends to be dynamic, e.g. legal form of a company might change through time, while the name does not vary. 

\begin{table}[H]
  \centering
  \begin{tabular}{>{\bfseries}l p{10cm}} 
    \toprule
    \textbf{Field Name} & \textbf{Description} \\
    \midrule
    Country of Residence & Indicates the country where the entity is based. \\
    Social Denomination & The official name of the entity. This column lists the full names of the companies or organizations. \\
    Entity/Application & A unique identifier or code assigned to the declaring entity within the dataset. \\
    National Identifier & A national identification number, likely a tax identification number or a similar unique identifier for the entity. Each country has its own and unique identification number. \\
    Identifier Type & Specifies the type of national identifier used. Some countries might present multiple types of identifier. \\
    Legal Entity Identifier (LEI) & A unique global identifier for legal entities participating in financial transactions. \\
    Sector & The industry or sector in which the entity operates. \\
    Legal Form & The legal structure of the entity. \\
    Abbreviation & An abbreviation or short form of the entity's legal form. \\
    \bottomrule
  \end{tabular}
  \caption{Description of Fields in the Dataset}
  \label{tab:field_descriptions}
\end{table}

\subsection{\index{Methods}Methods}\label{sec:Methods}

\subsubsection{\index{Distance methods}Distance Methods}\label{sec:distance_methods}

\textbf{Levenshtein Distance}, introduced by Vladimir Levenshtein in 1965, is a fundamental algorithm in the field of string matching and comparison. It quantifies the dissimilarity between two strings by calculating the minimum number of single-character edits required to transform one string into another. These modifications encompass the addition, removal, and replacement of characters. The algorithm's straightforwardness and efficiency have established it as a fundamental component in numerous applications, such as spell checking, DNA sequence alignment, and natural language processing (\cite{yacham2024levenshtein}). 

For instance, the Levenshtein distance between “kitten” and “sitting” is 3, as a minimum of three edits are necessary to transform one word into the other. These edits include:
\begin{enumerate}
    \item kitten → sitten (substitution of “s” for “k”)
    \item sitten → sittin (substitution of “i” for “e”)
    \item sittin → sitting (insertion of “g” at the end)
\end{enumerate}
An “edit” is characterized by one of three operations: inserting a character, deleting a character, or replacing a character (\cite{nam2019levenshtein}).

To run this distance we use the Python library \href{https://pypi.org/project/python-Levenshtein/}{Levenshtein}.\\

Among the existing metrics, \textbf{Cosine Distance} (which measures the angle between two vectors) is one of the most widely used. It is efficiently computed as the dot product of two normalized vectors (\cite{li2013distance}). This metric, is widely used in text comparison tasks, such as document similarity and information retrieval.\\
In certain cases, we normalize vectors by dividing them by their magnitude, resulting in a unit vector with a length of 1. To create a unit vector from a given vector a , we divide it by its magnitude, denoted as |a |. When working with unit vectors, the dot product is equivalent to the cosine of the angle between them. The cosine value typically ranges from 1 for vectors aligned in the same direction, through 0 for perpendicular vectors, to -1 for vectors pointing in opposite directions. However, since raw frequency values are non-negative, the cosine similarity for these vectors is restricted to the range 0–1 (\cite{jurafsky2025cosine}).
\begin{figure}[H]
        \centering
        \includegraphics[width=1\linewidth]{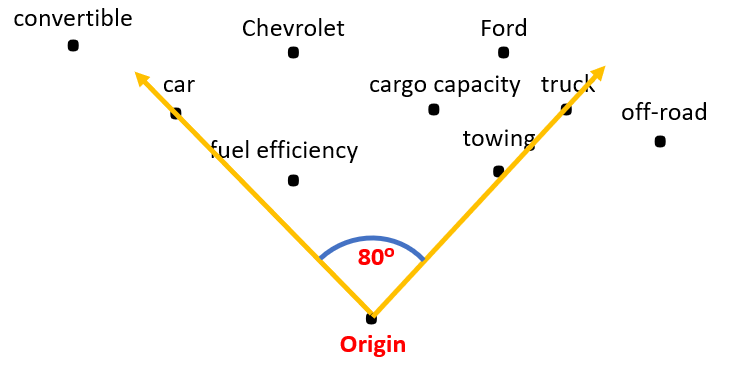}
        \caption{This example shows how cosine similarity will compare the angle of lines between objects to determine how similar the items are. Note that most text embeddings will be at least a few hundred dimensions instead of just two. Source: \cite{wiki:xxx}}
        \label{fig:cosinesimilarity}
    \end{figure}
Its computational efficiency makes it ideal for large-scale applications like search engines and recommendation systems. In Python, cosine distance can be computed using libraries like \href{https://pypi.org/project/scikit-learn/}{scikit-learn}.\\

The \textbf{Jaccard Distance} evaluates the similarity between two datasets by comparing their shared and unique elements. The Jaccard similarity is determined by dividing the count of common elements in both sets by the total number of elements present in either set. Essentially, the Jaccard similarity is calculated as the ratio of the intersection's size to the union's size of the two sets (\cite{karabiber2015jaccard}). The Jaccard similarity is calculated as:
\[
J(A, B) = \frac{|A \cap B|}{|A \cup B|}
\]
The Jaccard distance is then:
\[
D_{\text{Jaccard}}(A, B) = 1 - J(A, B)
\]
For example, comparing the sets {"cat", "dog", "mouse"} and {"cat", "bird"}, the intersection contains one element ("cat"), and the union contains four unique elements. The Jaccard similarity is $\frac{1}{4}$, so the distance is $\frac{3}{4}$. 

Jaccard Similarity serves as an effective method for measuring the resemblance between two sets, finding utility in a wide range of data analysis tasks. Whether applied to text mining, recommendation engines, or genomic research, leveraging this metric can significantly improve your capacity to extract valuable insights from complex datasets (\cite{zhou_jaccard_2023}).

In Python, it can be computed using libraries like \href{https://pypi.org/project/scipy/}{scipy} or custom implementations.

\subsubsection{\index{Large Language Models through huggingface}Large Language Models through huggingface}\label{sec:llm}

\textbf{BART (Bidirectional and Auto-Regressive Transformer)}
is an enhanced version of BERT, designed with a focus on generating natural text. It features a bidirectional encoder paired with an auto-regressive decoder. During training, the input text is first perturbed through various noise-inducing techniques, such as deletions and masking, after which the model attempts to reconstruct the original text in a sequential manner. This architecture has proven effective for numerous applications, including Machine Translation (MT), Machine Summarization (MS), and Question Answering (QA), making BART a versatile tool in natural language processing tasks (\cite{math10213967}).

The BART model's flexible architecture makes it suitable for a wide range of downstream tasks. For sequence classification, the final hidden state of the decoder’s last token is utilized, akin to how BERT employs its CLS token, but with the advantage of attending to the entire input sequence. In token classification tasks, such as those required by SQuAD, the top hidden states from the decoder are used to classify each token individually. BART’s autoregressive decoder is particularly effective for sequence generation tasks like abstractive question answering and summarization, which align seamlessly with its denoising pre-training objective. Furthermore, BART can be adapted for machine translation by incorporating a new encoder that maps foreign-language inputs into representations that BART can denoise into English, with the entire system fine-tuned end-to-end using cross-entropy loss (\cite{lewis2019bart}). To use BART, you can leverage libraries like Hugging Face's transformers in Python.\\

\textbf{BERT} which stands for Bidirectional Encoder Representations from Transformers, is a language model designed to capture deep bidirectional context by leveraging both the left and right context during pretraining on unlabeled text. This unique approach allows BERT to generate rich, context-aware representations that can be fine-tuned with minimal adjustments—often just an additional output layer—to achieve state-of-the-art performance across various tasks, including question answering and language inference, without requiring significant changes to task-specific architectures (\cite{devlin2019bertpretrainingdeepbidirectional}).

BERT is versatile for various natural language processing tasks. For classification tasks like sentiment analysis, a classification layer is added on top of the transformer's output corresponding to the [CLS] token. In question-answering tasks, where the goal is to identify the answer within a given text sequence, BERT can be fine-tuned by learning two additional vectors that indicate the start and end positions of the answer. Similarly, for Named Entity Recognition (NER), BERT can be utilized by passing the output vector of each token through a classification layer that predicts the appropriate NER label (\cite{shushanta_bert_2023}).\\

\textbf{DeBERTa} stands for decoding-enhanced BERT with disentangled attention. It improves the BERT and RoBERTa models using DA (disentangled attention) mechanism and enhanced mask decoder (\cite{assiri2024deberta}; \cite{aziz2022enhancing}). Unlike traditional methods that rely on a single vector to jointly represent the content and position of each input word, the Disentangled Attention (DA) mechanism employs two distinct vectors: one dedicated to capturing the content and another for the position. The attention weights between words are then calculated using disentangled matrices that separately account for both their content and relative positions. Similar to BERT, DeBERTa is pre-trained using a masked language modeling (MLM) objective. While the DA mechanism effectively incorporates the content and relative positions of context words, it does not consider their absolute positions, which are often critical for accurate predictions. To address this, DeBERTa introduces an enhanced mask decoder that augments the MLM process by incorporating absolute position information of the context words at the decoding layer (\cite{he2023debertav3improvingdebertausing}).

\subsubsection{\index{LLMs through interface}LLMs through interface}\label{sec:llm}

\textbf{Mistral}
has a chatbot service, similar to OpenAI’s ChatGPT, first released to beta on 26 February, 2024. Alongside Mistral Large and Mistral Small, Mistral recently added the multimodal Pixtral 12B to the roster of LLMs available in Le Chat(\cite{ibm_mistral_ai}).

Mistral AI offers a range of AI models, both commercial and open-source, tailored for diverse applications such as text generation, natural language processing (NLP), code development, and complex reasoning tasks. These models are engineered to be lightweight, efficient, and scalable, ensuring they consume fewer computational resources without compromising accuracy. By releasing fully open-source models, Mistral AI empowers organizations to adapt and deploy AI solutions that fit their specific requirements. The models support multiple languages and programming frameworks, enhancing their versatility for developers and researchers globally. With a focus on efficiency, transparency, and accessibility, Mistral AI distinguishes itself as a competitive alternative to industry leaders like OpenAI, Google DeepMind, and Anthropic. Their offerings deliver state-of-the-art performance while maintaining cost-effectiveness, making them an appealing option for businesses across various sectors (\cite{voiceflow_mistral_ai}).\\

\textbf{Microsoft Copilot} 
is integrated into Microsoft 365 in two ways. It works alongside you, embedded in the Microsoft 365 apps you use every day — Word, Excel, PowerPoint, Outlook, Teams and more — to unleash creativity, unlock productivity and up level skills. Today we’re also announcing an entirely new experience: Business Chat. Business Chat works across the LLM, the Microsoft 365 apps, and your data — your calendar, emails, chats, documents, meetings and contacts — to do things you’ve never been able to do before. You can give it natural language prompts like “Tell my team how we updated the product strategy,” and it will generate a status update based on the morning’s meetings, emails and chat threads (\cite{microsoft_copilot}).

Copilot gradually learns and adjusts to the user's writing style and preferences, offering increasingly customized suggestions as it becomes more familiar with their habits. Additionally, it can adapt its tone and content depending on the context, such as crafting a formal email for an important client versus a relaxed response for a colleague, ensuring the output aligns with the intended audience and purpose (\cite{microsoft_copilot_advantages}).\\

\textbf{Qwen}
is the LLM and large multimodal model series of the Qwen Team, Alibaba Group. Now the LLMs have been upgraded to Qwen2.5. Both language models and multimodal models are pretrained on large-scale multilingual and multimodal data and post-trained on quality data for aligning to human preferences. Qwen is capable of natural language understanding, text generation, vision understanding, audio understanding, tool use, role play, playing as AI agent, etc (\cite{QwenDocs}). 

Qwen AI is Alibaba’s answer to the AI race, competing with ChatGPT, Gemini, and DeepSeek AI. It excels in coding, long-text processing, and multilingual tasks, making it a strong choice for developers, businesses, and researchers (\cite{Qweneverything}).

\subsubsection{\index{Zero-shot Classification}Zero-shot Classification}\label{sec:Data_Methods}

After reviewing the different techniques and methods available, we decided to explore the application of zero-shot classification (ZSC) with LLMs in entity comparison processes. In zero-shot learning (ZSL), the model tackles a task without being exposed to any labeled data specific to that task. Instead, it leverages the knowledge it has acquired during its training on other tasks or its broad understanding of language and concepts to make predictions or decisions. This approach highlights the model's ability to generalize and apply learned patterns to new, unseen scenarios without explicit guidance or fine-tuning for the specific task at hand (\cite{nicoomanesh_fewshot_2024}). ZSC focuses on assigning a suitable label to a given text without being constrained by the text's domain or the specific aspect (such as topic, emotion, event, etc.) that the label represents. This approach leverages the model's general understanding and reasoning capabilities to make accurate predictions across diverse contexts without prior exposure to labeled examples in the target domain (\cite{yin2019benchmarkingzeroshottextclassification}).
Regarding the labels of this method, we will be working with 2 labels: \textit{Different} or \textit{Equal}.
The main reason behind this choice is linked to the absence of GPU\footnote{\href{https://community.fs.com/es/article/a-brief-introduction-to-cpu-gpu-asic-and-fpga.html}{Graphics Processing Unit: Designed to efficiently process and display graphics}} on our computers, in contrast we had to implement a method that used only the CPU\footnote{\href{https://community.fs.com/es/article/a-brief-introduction-to-cpu-gpu-asic-and-fpga.html}{Central Processing Unit: The brain of the computer, executing instructions and processing data for various tasks}}. This approach uses the attributes of LLMs to improve the accuracy and efficiency of EM processes. Traditionally, entity comparison was carried out manually. These methods were not only time-consuming but also prone to errors. The application of ZSC with LLMs can significantly ameliorate this process.

A common limitation in EM tasks is the reliance on labeled examples for training. However, a more flexible but challenging scenario is zero-shot entity matching (ZSEM), where the system must operate on an entirely unseen dataset without access to any labeled examples (\cite{zhang2024anymatchefficientzeroshot}).
Although currently our study does not dive into ZSEM, we understand its great potential for future research. ZSEM represents a flexible yet challenging approach, enabling entity comparison using LLMs without the need for labeled examples. This function could be particularly advantageous in scenarios where labeled data is insufficient or unavailable. By exploring ZSEM in future studies, we could improve our EM processes, making them more robust and adaptable to diverse datasets. We believe that integrating ZSEM into our methodology can lead to more effective results, and it adds significant value to our entity comparison processes.

\subsubsection{\index{Web Scraping}Web Scraping}\label{sec:webscraping}

After obtaining our dataset in an excel format, as aforementioned, we need to compare this information with the one stated in the official websites. To do so we will carry out some web-scraping methods.

The result of these methods will be joined to the initial dataset, so that our LLM can compare both names (declared, official).

The columns will be the next:
\begin{enumerate}
    \item Official: This column will contain the official names, established in the country's register.
    \item Previous names: If this entity has previous names, they will be shown in this column.
\end{enumerate}

\subsubsection{\index{Model Fine-Tuning}Model Fine-Tuning}\label{sec:Data_Methods}

In order to get the model to predict the correct label when classifying, we will have to fine-tune it with examples related to our data.
The fine-tuning of LLMs stands out as one of the most resource-intensive actions in machine learning, both in terms of data and computational demands. These models gain outstanding capabilities by training extensive transformer-based architectures on loads of data comprising trillions of tokens of text. This process enables them to learn complex patterns, contextual relationships, and generalizable knowledge, making them highly effective across a wide range of NLP\footnote{Natural Language Processing} tasks (\cite{sachdeva2024traindataefficientllms}). The training of LLMs can generally be categorized into three key stages. The first stage focuses on data collection and processing, where raw data is gathered, cleaned, and prepared for training. The second stage involves the pre-training process, during which the model's architecture is defined and pre-training tasks are established. Finally, the third stage consists of fine-tuning and alignment, where the model is adapted to specific tasks through supervised fine-tuning. Alignment techniques are applied to ensure the model's outputs are in tune with human preferences or desired behaviors (\cite{LIU2025129190}).\\
To fine-tune our LLMs and having in mind the information explained before, we followed these steps:
\begin{itemize}
    \item Firstly, we created a JSON\footnote{\href{https://en.wikipedia.org/wiki/JSON}{JavaScript Object Notation}} file with some examples, so the model understands what I need him to do. This file is composed of 433 observations. Structure of JSON file:
    \begin{figure}[H]
        \centering
        \includegraphics[width=1\linewidth]{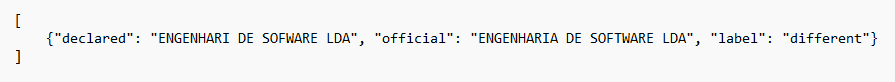}
        \caption{Example of the JSON file format}
        \label{fig:JSON_example}
    \end{figure}
    \item We analyze the class distribution before splitting the dataset to see if we need to over-sample the minority class. In an oversampling method, new samples are added to the minority class in order to balance the data set. These methods can be categorized into random oversampling and synthetic oversampling. In random oversampling method, the goal is to balance a dataset by increasing the representation of the minority class. This can be achieved through two main strategies: random oversampling and synthetic oversampling. Random oversampling works by duplicating existing examples from the minority class, effectively expanding its presence in the dataset. On the other hand, synthetic oversampling involves creating new, artificial examples for the minority class, which are designed to capture important patterns and reduce the risk of misclassifications. By enriching the minority class with these additional samples, the model becomes better equipped to learn its characteristics and make more accurate predictions (\cite{shelke2017review}). \\If needed, we duplicate the instances of the minority class, not only in train but also in test.
    \item Now it is time to load the model as well as the tokenizer.
    \item In this point, we can start with the training arguments. In this prompt, we include arguments like the learning rate, the size of the training and testing batch, the number of epochs and the metrics necessary to evaluate the model performance. \\In order to ensure a better understanding of the training dataset, we will conduct an Ensemble Prediction Function, to combine the predictions from multiple models.
    \item After having trained our model, we will pay attention to the metrics displayed to know if the model acted accordingly to our goal. \\To be sure that the model learns from the errors, we will carry out an error analysis so the model does not stumble over the same stone twice.
    \item Finally, we save the new trained model in a different path to continue with the fine-tuning later in the process.
\end{itemize}

\section{\index{Case Study: Portugal}Case Study: Portugal}\label{sec:use_case}

This case study is based on a real-world dataset collected in Portugal over a single day (see Table \ref{tab:validation_dataset} in the Appendix), specifically designed to evaluate the performance of LLMs and their interfaces in identifying and resolving EM process. This Dataset, serves as a valuable validation tool for assessing both the accuracy of responses generated by LLMs and the effectiveness of user interfaces used to interact with these models, as well as the aforementioned distance algorithms.

\subsection{\index{Name Comparison}Name Comparison}\label{sec:use_case}
In order to convey the name comparison, our dataset includes two additional columns that provide crucial information for analysis. The first column contains the officially verified names gathered through web scraping. The column "Result" acts as a ground truth against which the LLM-generated responses are compared. It categorizes each observation into one of three categories: "Aceptado" (Accepted), "Rechazado" (Rejected) or "Duda" (Not sure). These categories enable a systematic evaluation of the quality of responses generated by the models.

The "Result" column (see Table \ref{tab:validation_dataset}) plays a pivotal role in the evaluation of the model. Responses classified as "Accepted" correspond to those where the entity provided a correct output, compared to the officially recorded name. In contrast, responses marked as "Rejected" indicate an error committed by the entity. These errors range from bad interpretation to spelling inconsistencies, where model abilities are needed to ensure high precision and contextual understanding. Finally, responses labeled as "Doubtful" represent ambiguous or complex cases requiring domain knowledge carried out by a human specialist.

\subsection{\index{Legal Form Comparison}Legal Form Comparison}\label{sec:use_case}
In our analysis of the legal forms of entities, we followed a multi-step process to ensure accuracy. Firstly, we separated the legal form of the reporting entity from both the declared and the official name. This allowed us to independently verify the consistency of the legal forms associated with each entity. After the first comparison is done and the resolution is positive, we move on to compare these two legal forms with the one declared in the legal form field of our dataset (see Table \ref{tab:legalform_dataset}). These two comparisons enable us to identify any discrepancies between the reported legal form and the actual legal form of the entity. By cautiously analyzing these differences, we are able to draw well-founded conclusions regarding the accuracy and reliability of the legal forms.

\begin{table}[h]
    \centering
    \begin{threeparttable}
    \scriptsize
    \raggedright 
    \begin{tabular}{cclccc}
        \hline
        \textbf{LF\tnote{a}} & \textbf{Abbreviation} & \textbf{Official Name} & \textbf{Result} & \textbf{Declared LF} & \textbf{Official LF}\\
        \hline
         &  & \\
        PTXXX & LDA\tnote{b} & \makecell[l]{BRENNTAG PORUTGAL - \\ PRODUTOS QUIMICOS, \\ LDA} & Accepted & LTDA & LDA\\
         &  & \\
        PTXXX & SA\tnote{c} & SILVER HORSE, S.A. & Accepted & SA & SA\\
         &  & \\
        PTXXX & LDA & \makecell[l]{VMAXPARTS, \\ UNIPESSOAL, \\ LDA} & Accepted & UNIPESSOAL & UNIPESSOAL \\
        \hline
    \end{tabular}
    \begin{tablenotes}
        \item[a] Legal Form
        \item[b] Private limited company (Ltd)
        \item[c] Public limited company (PLC)
    \end{tablenotes}
    \end{threeparttable}
    \caption{Legal Forms}
    \label{tab:legalform_dataset}
\end{table}

\section{\index{Results}Results}\label{sec:Results}

In evaluating the performance of these methods, several key metrics were considered: Accuracy, Precision, Recall, F1 Score, ROC AUC, and False Positive Rate (FPR). Accuracy measures the proportion of correct predictions over total predictions, providing an overall assessment of model effectiveness. Precision quantifies the proportion of correctly predicted positive instances out of all instances that were predicted as positive, answering the question: How reliable are the model’s positive predictions? 

Recall evaluates the model’s ability to identify all relevant cases, which can be important in certain scenarios; however, the impact of failing to detect a positive case must be weighed alongside other metrics. The F1 Score strikes a balance between precision and recall, though its relevance is relatively diminished in situations where the consequences of specific errors are more critical. ROC AUC estimates the model’s ability to distinguish between classes, with higher values indicating superior discriminative power. Nevertheless, in our context, the primary focus was placed on the False Positive Rate (FPR), calculated as the proportion of negative observations incorrectly classified as positive. This metric is crucial for measuring the risk of erroneously accepting incorrect entities—a scenario that is unacceptable in our case. Given the critical implications of minimizing such errors, we prioritized analyzing the FPR to ensure that incorrect entities were reliably rejected, even if this came at the expense of performance in other metrics.

The results obtained from our analysis reveal remarkable variations in performance across different methods. Traditional distance-based methods demonstrated robust performance, with Levenshtein achieving an accuracy of 92.06\%, precision of 96.49\%, recall of 94.83\%, and an F1 Score of 95.65\%. However, its FPR stood at 40\%, indicating room for improvement in minimizing false positives. Cosine similarity showed even stronger results, with accuracy reaching 93.65\%, precision at 98.21\%, and F1 Score of 96.49\%, while maintaining a significantly lower FPR of 20\%. Among the Huggingface LLM models, deBERTa-v3-base stood out with perfect recall and a high F1 Score of 95.87\%, though its ROC AUC score was moderate at 50\%—indicating performance no better than random chance—and its FPR was relatively high at 100\%, suggesting challenges in avoiding false positives despite its excellent recall. Notably, the LLM Interface models, particularly Microsoft/Copilot, achieved exceptional performance with accuracy of 93.65\%, perfect recall, F1 Score of 96.67\%, and ROC AUC of 60\%, in contrast with an FPR of 80\%. Similarly, Alibaba's Qwen2.5 model exhibited strong performance with accuracy at 95.24\%, precision of 96.61\%, and F1 Score of 97.44\%, while maintaining a more acceptable FPR of 40\% (\cite{aydin2025generative}). These findings underscore the superior capabilities of interface-based LLMs in complex classification tasks, probably due to their extensive training on diverse datasets and advanced architectures that enable them to capture complex contextual relationships.

In contrast, some models like Facebook/BART-large-mnli showed limitations, with a relatively low ROC AUC score of 44.83\% and an FPR of 40\%, suggesting challenges in distinguishing between classes effectively. This discrepancy can be attributed to the inherent differences in model architectures and training methodologies, where certain models may excel in specific tasks but struggle in others due to less effective fine-tuning or insufficient adaptation to particular domains (\cite{lewis2019bart}). 

In general, LLMs tend to give more variate answers when there happens to be abreviations or lexic variations in the names. For instance, PASTIGEST IND E COM PASTELARIA DOCARIA BISCOIT E GELADOS SA is the declared name and PASTIGEST - INDÚSTRIA E COMÉRCIO DE PASTELARIA, DOÇARIA, BISCOITOS E GELADOS, S.A. is the official name. We can see abreviations like IND for INDUSTRIA or COM for COMERCIO. These inconsistencies seem challenging for the models due to the lack of specific fine-tuning examples that capture complex patterns, particularly in financial or legal contexts.

Overall, our study highlights the potential of LLMs in improving EM processes. By leveraging their ability to process and interpret diverse linguistic variations, LLMs offer a promising path for improving the accuracy and efficiency of EM in financial systems. The substantial differences in performance metrics across models emphasize the need for careful selection of different approaches to achieve optimal results. While traditional distance-based methods remain effective for simple matching tasks, LLM models provide better performance in complex classification scenarios, demonstrating their capability to minimize human intervention in critical decision-making processes. However, given the restriction of zero tolerance for accepting incorrect entities when handling legal and confidential data, we will focus on refining these models further, expanding their knowledge base, and ensuring consistent performance across diverse datasets. Special attention will be paid to reducing the FPR to eliminate the risk of erroneously classifying incorrect entities as valid, ultimately leading to more reliable and scalable solutions worldwide.

\begin{sidewaystable}[!htbp] \centering \small
  \centering
    \begin{tabular}{lcccccc}
     \midrule 
     {\textbf{Method}}
        & \multicolumn{1}{c}{\textbf{Accuracy}} 
        & \multicolumn{1}{c}{\textbf{Precision}} 
        & \multicolumn{1}{c}{\textbf{Recall}} 
        & \multicolumn{1}{c}{\textbf{F1 Score}}
        & \multicolumn{1}{c}{\textbf{ROC AUC}}
        & \multicolumn{1}{c}{\textbf{FPR\footnote{False Positive Rate}}} \\
     \midrule  
     \textbf{Distance}\\
\vspace{-0.5em} & & & & & \\ 
       Levenshtein & 92.06 & 96.49& 94.83 & 95.65 & 77.41 & 40\\
       Cosine & 93.65 & \textbf{98.21} & 94.83 & 96.49 & \textbf{87.41} & \textbf{20}\\
       Jaccard & 58.73 & 97.06 & 56.90 & 71.74 & 68.45 & 80\\
\midrule
\textbf{LLM Huggingfce}\\
\vspace{-0.5em} & & & & & \\ 
       Facebook/BART-large-mnli\footnote{\href{https://huggingface.co/facebook/bart-large-mnli}{Bart-large after being trained on the Multi-NLI (MNLI) dataset.}} & 82.54 & 91.23 & 89.66 & 90.43 & 44.83 & 40\\
       deBERTa-v3-base\footnote{\href{https://huggingface.co/microsoft/deberta-v3-base}{DeBERTa improves the BERT and RoBERTa models using disentangled attention and enhanced mask decoder.}} & 92.06 & 92.06 & \textbf{100.00} & 95.87 & 50.00 & 100\\
       BERT-base-uncased\footnote{\href{https://huggingface.co/docs/transformers/model_doc/bert}{BERT, is a bidirectional transformer pretrained using a combination of masked language modeling objective and next sentence prediction on a large corpus comprising the Toronto Book Corpus and Wikipedia.}} & 82.54 & 91.23 & 89.66 & 90.43 & 44.83 & 100\\
       \midrule 
     \textbf{LLM Interface}\\
 \vspace{-0.5em} & & & & & \\ 
       Alibaba/Qwen2.5\footnote{\href{https://chat.qwenlm.ai/}{Qwen2.5 is the latest series of Qwen large language models. For Qwen2.5, we release a number of base language models and instruction-tuned language models ranging from 0.5 to 72 billion parameters.}} & \textbf{95.24} & 96.61 & 98.28 & \textbf{97.44} & 79.14 & 40\\
       MISTRAL AI SAS/Mistral\footnote{\href{https://mistral.ai/}{Mistral AI is a research lab building the best open source models in the world.}} &  92.06  & 93.44 &  98.28 & 95.80 & 59.14 & 80\\
       Microsoft/Copilot\footnote{\href{https://copilot.microsoft.com}{Microsoft 365 Copilot is an AI-powered productivity tool that uses large language models (LLMs) and integrates your data with the Microsoft Graph and Microsoft 365 apps and services.}} & 93.65 & 93.55 &  \textbf{100.00} & 96.67 & 60.00 & 80\\
       \midrule 
     \end{tabular}
     \label{tab:results}
  \end{sidewaystable}

\section{\index{Conclusions and further research}Conclusions and Further Research}\label{sec:Conclusion}

This study has demonstrated the successful application of LLMs compared to similarity algorithms (Levenshtein, Jaccard ...) in addressing EM challenges using a dataset like the one in the Table \ref{tab:validation_dataset} in the Appendix. This Dataset, was designed to evaluate the performance of LLMs including their interfaces and distance methods in entity comparison, served as a crucial validation tool for evaluating the accuracy of model-generated responses. By adding officially verified names as ground truth and categorizing results into "Accepted," "Rejected," or "Doubtful," we established a robust way for evaluating model performance and identifying areas for improvement.

The findings reveal substantial differences regarding the performance of the different methods used. Traditional distance-based algorithms, such as Levenshtein and Cosine similarity, appeared to be highly reliable in specific contexts, particularly in terms of accuracy and recall. Among the Huggingface models, deBERTa-v3-base model had a perfect recall and F1 scores, highlighting its robustness in identifying positive cases. Furthermore, LLM interface models like Copilot present surprising performance in all metrics, underlining their strong capabilities in complex classification tasks. These results shows up the potential of LLMs to enhance decision-making processes though minimizing human intervention in critical tasks.

Thinking about the near future, our main goal is to extrapolate and adapt these methodologies to all the possible countries. Based on the Portugal case study, we are seeking to apply the same framework to diverse countries. This augmentation will involve addressing specific challenges, including the change of data description and linguistic diversity, which may complicate model functions. Additionally, we plan to refine our evaluation metrics to ensure balanced and solid performance across all criteria, particularly for models that currently exhibit lower ROC AUC scores. The closer the ROC AUC is to 0.5, the worst. Having this ROC AUC score, means that the probability of choosing between the different labels is done randomly (for instance, throwing a coin in the air has the same probability of landing in the heads or the tails).

To sum up, while this study provides a solid foundation for the use of LLMs in particular cases, there are still many ways of gaining leverage of these in other contexts. Future research will focus on developing stronger fine-tuning, expanding their knowledge, and ensuring accurate capabilities across diverse countries, this will eventually lead to a more reliable analyses worldwide.

\newpage
\bibliographystyle{bde-en}
\bibliography{Bibliografia}
\newpage

\appendix

\section{Appendix – Additional figures and tables} \label{sec:Appendix}

\subsection{\index{Validation Dataset}Validation Dataset}\label{sec:Validation Dataset}


\begin{table}[h]
\begin{adjustbox}{width=1.2\textwidth, center}
\begin{threeparttable}
\centering
\small
    \begin{tabular}{ccccccccccc}
        \hline
        \textbf{Country} & \textbf{Company name} & \textbf{Entity} & \textbf{National Identifier} & \textbf{Identifier type} & \textbf{LEI\tnote{a}} & \textbf{Sector} & \textbf{Legal Form} & \textbf{Abbreviation} & \textbf{Official Name} & \textbf{Result} \\
        \hline
        & & \\
        PT & \makecell[l]{SOLARSHOP, \\ UNIPESSOAL LDA} & X & XXXX & PT\_NIF\_CD & & SXX & PTXXX & LDA & \makecell[l]{SOLARSHOP, \\ UNIPESSOAL, LDA} & Accepted \\
        & & \\
        PT & \makecell[l]{SIMBOLO II - \\ INFORMATICA DE \\ GESTAO LDA} & X &  &  & X & SXX & PTXXX & LDA &  & Doubtful \\
        & & \\
        PT & \makecell[l]{ALTRAD PREFAL, \\ UNIPESSOAL LDA} & X & XXXX & PT\_NIF\_CD & & SXX & PTXXX & LDA & \makecell[l]{LTRAD SERVICES \\ INDUSTRIE, UNIPESSOAL \\ LDA} & Rejected \\
        & & \\
        \hline
    \end{tabular}
\begin{tablenotes}
    \item[a] Legal Entity Identification
    \end{tablenotes}
\end{threeparttable}
    \label{tab:validation_dataset}
    \end{adjustbox}
\end{table}

\subsection{\index{Manual Comparison Process}Manual Comparison Process}\label{sec:Manual_Comparison_Process}

The current manual comparison process involves several meticulous steps to ensure the accurate identification and codification of foreign entities. This process is carried out by trained personnel who utilize various tools and resources to verify the data provided by the entities. The following steps outline the manual comparison process:
\begin{itemize}[label={$\bullet$}]
    \item \textbf{Accessing the Application:} The process begins by accessing the gnrmain application through the provided link. Users navigate to the "Peticiones de Personas Jurídicas - Entidades" section to view pending requests.
    \item \textbf{Filtering and Searching Requests:} Users can filter and search for specific requests using various criteria such as request date, entity name, and identification number. This helps in narrowing down the list of pending requests that need to be resolved.
    \item \textbf{Verifying Request Details:} Once a request is selected, users must verify the details provided in the request. This includes checking the entity's name, identification number, legal form, country of residence, and other relevant information. The verification process involves cross-referencing the provided data with external sources such as national registries, legal forms lists, and LEI databases.
    \item \textbf{Assigning Codes:} After verifying the details, users assign the appropriate code to the entity. This code is crucial for the identification and tracking of the entity within the Spanish financial system. The assigned code is then recorded in the system.
    \item \textbf{Rejecting Requests:} If the request contains errors or inconsistencies, it is rejected. Users can select from a list of rejection reasons or describe the reason in text. The system allows users to highlight the fields that contain errors, facilitating the rejection process.
    \item \textbf{Reprocessing Requests:} In cases where a request was incorrectly accepted or rejected, users can reprocess the request. This involves selecting the request from the pending list and reprocessing it to resolve any issues.
\end{itemize}
This manual process, while thorough, is time-consuming and prone to human errors. It requires significant resources and expertise to ensure accuracy and compliance with regulatory standards.

\subsection{\index{Legal Form Comparison Process}Legal Form Comparison Process}\label{sec:Data Comparison}

In our comprehensive analysis of the legal forms of entities, we implemented a multi-step approach to ensure accuracy and thoroughness. Initially, we separated the legal form of the reporting entity from both the declared name and the official name. This distinction was crucial as it allowed us to independently verify the consistency and accuracy of the legal forms associated with each entity. By isolating these elements, we could focus on the specific attributes of the legal form without the potential biases introduced by the names.

Subsequently, we compared these two legal forms—the one associated with the declared name and the one associated with the official name—with the legal form declared in the legal form field of our dataset. This step was essential to identify any discrepancies or inconsistencies between the reported legal form and the actual legal form of the entity. By conducting this comparison, we aimed to uncover any deviations that might indicate errors in reporting or differences in legal structuring.

It is important to note that the legal form field in our dataset follows a specific format: it consists of the initials of the country (e.g., PT for Portugal) followed by three numbers that vary according to the legal form. This standardized format facilitated our comparison process by providing a clear and consistent reference for each entity's legal form.

To ensure the robustness of our analysis, we employed a meticulous approach in examining these differences. We analyzed the nature and extent of any discrepancies, considering factors such as the jurisdictional variations in legal form definitions and the potential for misclassifications. This detailed examination provided us with a deeper understanding of the accuracy and reliability of the reported legal forms.

Based on our findings, we were able to draw well-founded conclusions regarding the legal structuring of the entities under review. Our methodical approach not only enhanced the robustness of our analysis but also provided valuable insights into the legal forms and their implications for the entities. This comprehensive analysis contributes to a better understanding of the legal landscape in which these companies operate and highlights the importance of accurate legal form reporting.

\subsubsection{\index{Distance methods}Distance Methods}\label{sec:distance_methods}
\begin{flushleft}

\textbf{Levenshtein Distance}
\end{flushleft}
The Levenshtein distance, also known as edit distance, is a measure of the difference between two sequences of characters. It is defined as the minimum number of operations needed to transform one sequence into another. The allowed operations are:
\begin{itemize}
    \item \textbf{Insertion}: Adding a character at any position in the sequence.
    \item \textbf{Deletion}: Removing a character from any position in the sequence.
    \item \textbf{Substitution}: Changing one character to another at any position in the sequence.
\end{itemize}

Mathematically, if we have two strings \(a\) and \(b\) of lengths \(m\) and \(n\) respectively, the Levenshtein distance \(d(a, b)\) can be calculated using a matrix \(D\) of size \((m+1) \times (n+1)\), where \(D[i][j]\) represents the distance between the first \(i\) characters of \(a\) and the first \(j\) characters of \(b\). The matrix is initialized as follows:
\[
D[i][0] = i \quad \text{for} \quad 0 \leq i \leq m
\]
\[
D[0][j] = j \quad \text{for} \quad 0 \leq j \leq n
\]

Then, the matrix is filled using the following recurrence relation:
\[
D[i][j] = \min \begin{cases} 
D[i-1][j] + 1 \\
D[i][j-1] + 1 \\
D[i-1][j-1] + \text{cost} 
\end{cases}
\]
where the cost is 0 if \(a[i-1] = b[j-1]\) and 1 otherwise. This algorithm has a time complexity of \(O(m \times n)\).
\vspace{3em} 
\begin{flushleft}
\textbf{Cosine Distance}
\end{flushleft}

The cosine distance is a measure of similarity between two vectors in a multidimensional space. It is primarily used in natural language processing and text mining to compare documents or phrases. The cosine similarity is defined as the cosine of the angle between two vectors \(A\) and \(B\), and is calculated as:
\[
\text{Cosine Similarity} = \frac{A \cdot B}{||A|| \times ||B||}
\]
where \(A \cdot B\) is the dot product of the vectors and \(||A||\) and \(||B||\) are the magnitudes of the vectors. The dot product is calculated as:
\[
A \cdot B = \sum_{i=1}^{n} A_i B_i
\]
and the magnitude of a vector \(A\) is calculated as:
\[
||A|| = \sqrt{\sum_{i=1}^{n} A_i^2}
\]

Cosine similarity ranges from -1 to 1, where 1 indicates that the vectors are identical, 0 indicates no similarity, and -1 indicates that they are opposite. This method is particularly useful when working with high-dimensional data, such as in information retrieval and product recommendation systems, as it is less sensitive to the magnitude of the vectors and focuses on the direction.

\vspace{3em} 
\begin{flushleft}
\textbf{Jaccard Distance}
\end{flushleft}

The Jaccard distance measures the dissimilarity between two sets by calculating the proportion of the intersection of the sets over their union. It is a binary similarity measure, useful for comparing sets of binary data. The formula for Jaccard distance is:
\[
\text{Jaccard Distance} = 1 - \frac{|A \cap B|}{|A \cup B|}
\]
where \( |A \cap B| \) is the number of elements common to both sets \(A\) and \(B\), and \( |A \cup B| \) is the total number of unique elements in both sets.

For example, if \(A\) and \(B\) are two sets of words from two documents, the Jaccard distance tells us how different they are in terms of content. The smaller the distance, the more similar the sets are. This method is useful in applications such as duplicate detection, document classification, and data mining, as it provides a clear measure of similarity between sets of binary data.

\subsubsection{\index{Large Language Models through huggingface}Large Language Models through huggingface}\label{sec:llm}

\begin{flushleft}
\textbf{Facebook/BART}
\end{flushleft}

BART (Bidirectional and Auto-Regressive Transformers) is a transformer model proposed by Lewis et al. in 2019. It combines the bidirectional encoder of BERT with the autoregressive decoder of GPT. BART is pre-trained by corrupting text with an arbitrary noising function and then learning to reconstruct the original text.

BART uses a standard sequence-to-sequence (seq2seq) architecture with a bidirectional encoder and a left-to-right autoregressive decoder. The encoder is similar to BERT, while the decoder is similar to GPT.

The pre-training involves two main steps:
\begin{enumerate}
    \item Corrupting the input text using various noising functions such as token masking, token deletion, and sentence permutation.
    \item Training the model to reconstruct the original text from the corrupted input.
\end{enumerate}

BART is particularly effective for text generation tasks such as summarization, translation, and question answering. It also performs well on comprehension tasks.

\vspace{3em}
\begin{flushleft}
\textbf{Google Research/BERT}
\end{flushleft}

BERT (Bidirectional Encoder Representations from Transformers) is a transformer model proposed by Devlin et al. in 2018. It is designed to pre-train deep bidirectional representations by jointly conditioning on both left and right context in all layers.

BERT uses a bidirectional transformer encoder. Unlike traditional left-to-right or right-to-left language models, BERT considers both directions simultaneously.

BERT is pre-trained using two tasks:
\begin{enumerate}
    \item \textbf{Masked Language Modeling (MLM)}: Randomly masking some of the tokens in the input and training the model to predict the masked tokens.
    \item \textbf{Next Sentence Prediction (NSP)}: Training the model to predict whether a given sentence is the next sentence in the original text.
\end{enumerate}

BERT is widely used for various natural language understanding tasks such as text classification, named entity recognition, and question answering.

\vspace{3em}
\begin{flushleft}
\textbf{Microsoft/deBERTa}
\end{flushleft}

DeBERTa (Decoding-enhanced BERT with Disentangled Attention) is a transformer model proposed by He et al. in 2021. It improves upon BERT and RoBERTa by using disentangled attention and an enhanced mask decoder.

DeBERTa uses a similar architecture to BERT but introduces two key improvements:
\begin{enumerate}
    \item \textbf{Disentangled Attention Mechanism}: Separates the attention into content and position information, allowing the model to better capture the relationships between words.
    \item \textbf{Enhanced Mask Decoder}: Improves the mask prediction by incorporating more context information.
\end{enumerate}

DeBERTa is pre-trained using a similar approach to BERT but with the additional disentangled attention mechanism and enhanced mask decoder.

DeBERTa achieves state-of-the-art performance on various natural language understanding tasks, including the GLUE benchmark and SQuAD.

\subsubsection{\index{Large Language Models through interface}Large Language Models through interface}\label{sec:llm}

\begin{flushleft}
\textbf{Alibaba/Qwen}
\end{flushleft}

Qwen is a dynamic AI platform designed to enhance interaction with advanced AI models. It supports both open-source and proprietary versions, offering a wide array of tools tailored for developers, researchers, and AI enthusiasts.

Qwen uses a transformer-based architecture with multiple versions, such as Qwen 2.5 Plus. The platform is built on the Open Web UI framework, prioritizing accessibility while introducing advanced features like artifact generation and model comparison.

The pre-training involves:
\begin{enumerate}
    \item \textbf{Data Cleaning and Balancing}: Ensuring high-quality and balanced datasets for training.
    \item \textbf{Multi-Task Learning}: Training the model on multiple tasks to improve generalization and performance.
\end{enumerate}

Qwen is ideal for complex coding tasks, including code generation, code repair, and reasoning across multiple programming languages. It is widely used in development environments to assist programmers.

\vspace{3em}
\begin{flushleft}
\textbf{MISTRAL AI SAS/Mistral}
\end{flushleft}

Mistral is a versatile AI platform that offers customizable models for various applications. It provides enterprise-grade AI solutions that can be deployed on-premises, in the cloud, or at the edge.

Mistral models are based on transformer architectures and are designed to be highly configurable. They support fine-tuning, distillation, and iteration to meet specific needs.

Mistral models are pre-trained using large-scale datasets and advanced training techniques to ensure high performance and adaptability. The pre-training process includes:
\begin{enumerate}
    \item \textbf{Large-Scale Data Collection}: Gathering diverse and extensive datasets.
    \item \textbf{Advanced Training Techniques}: Using techniques such as transfer learning and fine-tuning to enhance model performance.
\end{enumerate}

Mistral models are used in a wide range of applications, including natural language processing, computer vision, and conversational AI. They are particularly effective in enterprise environments for tasks such as customer support, data analysis, and automation.

\vspace{3em}
\begin{flushleft}
\textbf{Microsoft/Copilot}
\end{flushleft}

Microsoft Copilot is an AI companion designed to assist users with various tasks, providing advice, feedback, and straightforward answers. It integrates seamlessly with Microsoft 365 and other Microsoft products.

Copilot uses a transformer-based architecture, leveraging the latest advancements in large language models. It is designed to be user-friendly and highly interactive.

Copilot is pre-trained on a vast amount of data, including text from the web, documents, and other sources. The pre-training process involves:
\begin{enumerate}
    \item \textbf{Data Collection}: Aggregating diverse datasets to cover a wide range of topics.
    \item \textbf{Fine-Tuning}: Adjusting the model to improve its performance on specific tasks and user interactions.
\end{enumerate}

Copilot is used to enhance productivity and provide support in various contexts, such as writing assistance, information retrieval, and task automation. It is integrated into Microsoft 365 to help users with tasks like drafting emails, creating documents, and analyzing data.

\end{document}